\documentclass[10pt,twocolumn,letterpaper]{article}

 \usepackage[pagenumbers]{cvpr} %

\usepackage[dvipsnames]{xcolor}

\definecolor{cvprblue}{rgb}{0.21,0.49,0.74}
\usepackage[pagebackref,breaklinks,colorlinks,citecolor=cvprblue]{hyperref}

\usepackage{booktabs}
\usepackage{multirow}
\usepackage{multicol}
\usepackage{color, colortbl}
\usepackage{pifont}%
\usepackage{array}

\def\paperID{4350} %
\def\confName{CVPR}
\def\confYear{2024}

\title{GenAD: Generative End-to-End Autonomous Driving}

\author{Wenzhao Zheng$^{1,}$\footnotemark[1]\quad Ruiqi Song$^{2,3,}$\footnotemark[1] \quad Xianda Guo$^{2,}$\footnotemark[1]$^{\ ,}$\footnotemark[2]  \quad Chenming Zhang$^{2}$ \quad  Long Chen$^{2,3,}$\footnotemark[2] \\
$^{1}$University of California, Berkeley \ \ \ \  $^{2}$Waytous \\
$^{3}$Institute of Automation, Chinese Academy of Sciences \\
\texttt{\small \{wenzhao.zheng, chmzhang\}@outlook.com; xianda\_guo@163.com; \{ruiqi.song, long.chen\}@ia.ac.cn} \\
}

\begin{document}
\maketitle

\renewcommand{\thefootnote}{\fnsymbol{footnote}}
\footnotetext{$\ast$ Equal contributions; $\dag$ Corresponding authors.}
\renewcommand{\thefootnote}{\arabic{footnote}}

\begin{abstract}
Directly producing planning results from raw sensors has been a long-desired solution for autonomous driving and has attracted increasing attention recently.
Most existing end-to-end autonomous driving methods factorize this problem into perception, motion prediction, and planning.
However, we argue that the conventional progressive pipeline still cannot comprehensively model the entire traffic evolution process, e.g., the future interaction between the ego car and other traffic participants and the structural trajectory prior. 
In this paper, we explore a new paradigm for end-to-end autonomous driving, where the key is to predict how the ego car and the surroundings evolve given past scenes.
We propose GenAD, a generative framework that casts autonomous driving into a generative modeling problem.
We propose an instance-centric scene tokenizer that first transforms the surrounding scenes into map-aware instance tokens.
We then employ a variational autoencoder to learn the future trajectory distribution in a structural latent space for trajectory prior modeling.
We further adopt a temporal model to capture the agent and ego movements in the latent space to generate more effective future trajectories.
GenAD finally simultaneously performs motion prediction and planning by sampling distributions in the learned structural latent space conditioned on the instance tokens and using the learned temporal model to generate futures.
Extensive experiments on the widely used nuScenes benchmark show that the proposed GenAD achieves state-of-the-art performance on vision-centric end-to-end autonomous driving with high efficiency.
Code: \url{https://github.com/wzzheng/GenAD}.

\end{abstract}

\section{Introduction}
\label{sec:intro}
Vision-centric autonomous driving has been extensively explored in recent years due to its economic convenience~\cite{caddn,bevdepth,lss,bevdet,bevformer}.
While researchers have advanced the limit of vision-centric autonomous driving in various tasks including 3D object detection~\cite{bevdepth,bevdet,bevformer}, map segmentation~\cite{liao2022maptr,liu2022vectormapnet,li2022hdmapnet,beverse}, and 3D semantic occupancy prediction~\cite{tpvformer,surroundocc,tian2023occ3d,tong2023scene,openoccupancy,pointocc,selfocc}, recent advances in vision-centric end-to-end autonomous driving~\cite{hu2023uniad,hu2022stp3,vad} have shed light on a potential and elegant path to directly produce planning results from raw sensors.

\begin{figure}[t]
\centering
\includegraphics[width=0.475\textwidth]{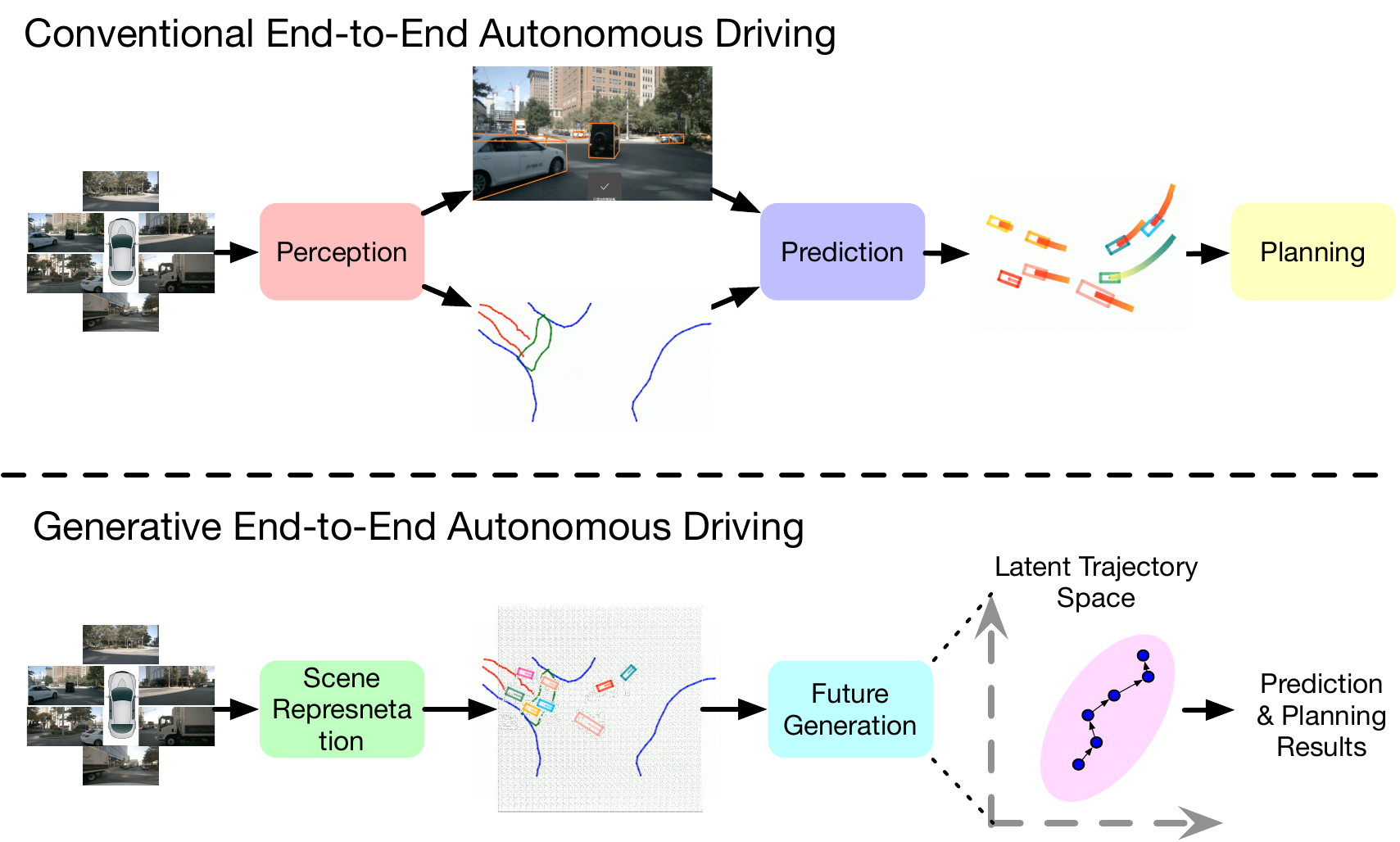}
\vspace{-7mm}
\caption{\textbf{Comparisons of the proposed generative end-to-end autonomous driving framework with the conventional pipeline.}
Most existing methods follow a serial design of perception, prediction, and planning.
They usually ignore the high-level interactions between the ego car and other agents and the structural prior of realistic trajectories.
We model autonomous driving as a future generation problem and conduct motion prediction and ego planning simultaneously in a structural latent trajectory space.
}
\label{fig:comparison}
\vspace{-6mm}
\end{figure}

Most existing end-to-end autonomous driving models are composed of several modules and follow a pipeline of perception, motion prediction, and planning~\cite{hu2023uniad,hu2022stp3,vad,ppad}. 
For example, UniAD~\cite{hu2023uniad} further progressively performs map perception, detection, tracking, motion prediction, occupancy prediction, and planning modules to improve the robustness of the system. 
It is also observed that using a planning objective improves the performance of intermediate tasks~\cite{hu2023uniad}.
However, the serial design of prediction and planning of existing pipelines ignores the possible future interactions between the ego car and the other traffic participants.
We argue that this type of interaction is important for accurate planning.
For example, the lane shift of the ego car would affect the action of the rear cars, and further affects the planning of the ego car.
This high-order interaction cannot be effectively modeled by the current design of motion prediction before planning.
Also, future trajectories are highly structured and share a common prior (e.g., most trajectories are continuous and straight lines).
Still, most existing methods fail to consider this structural prior, leading to inaccurate predictions and planning.

In this paper, we propose a \textbf{Gen}erative End-to-End \textbf{A}utonomous \textbf{D}riving (GenAD) framework (shown in Figure~\ref{fig:comparison}), which models autonomous driving as a trajectory generation problem to unleash the full potential of end-to-end methods.
We propose a scene tokenizer to obtain instance-centric scene representations, which focus on instances but also integrate map information.
To achieve this, we use a backbone network to extract image features for each surrounding camera and then transform them into the 3D bird's eye view (BEV) space~\cite{bevdet,bevformer,bevdepth}.
We further use cross-attention to refine high-level map and agent tokens from BEV features.
We then add an ego token and use ego-agent self-attention to capture their high-order interactions.
We further inject map information with cross-attention to obtain map-aware instance tokens.
To model the structural prior of future trajectories, we learn a variational autoencoder to map ground-truth trajectories to Gaussian distributions considering the uncertain nature of motion prediction and driving planning.
We then use a simple yet effective gated recurrent unit (GRU)~\cite{gru} to perform auto-regressing to model instance movement in the latent structural space.
During inference, we sample from the learned distributions conditioned on the instance-centric scene representation and can thus predict different possible futures.
Our GenAD can simultaneously perform motion prediction and planning using the unified future trajectory generation model.
We conduct extensive experiments on the widely used nuScenes benchmark to evaluate the performance of the proposed GenAD framework. 
Based on generative modeling, our GenAD achieves state-of-the-art vision-based planning performance with high-efficiency.

\section{Related Work}
\label{sec:formatting}

\textbf{Perception.} 
Perception is the basic step in autonomous driving, which aims to extract meaningful information from raw sensor inputs.
Despite the strong performance of LiDAR-based methods~\cite{voxelnet, voxelnext, voxeltr, sparseconv}, vision-centric methods~\cite{tpvformer, beverse,surroundocc,bevdepth,bevformer} have emerged as a competitive alternative due to the low costs of RGB cameras.
Equipped with large 2D image backbones, vision-centric methods demonstrated great potential in the main perception tasks including 3D object detection~\cite{caddn,bevdepth,lss,bevdet,beverse,bevformer}, HD map reconstruction~\cite{liao2022maptr,liu2022vectormapnet,li2022hdmapnet,beverse}, and 3D semantic occupancy prediction~\cite{surroundocc,tian2023occ3d,tong2023scene,openoccupancy,pointocc,occworld}.
To accurately complete these 3D tasks, the key procedure is to transform image features to the 3D space.
One line of works predicts explicit depths for image features and then projects them into the 3D space using camera parameters~\cite{caddn,bevdepth,bevfusion,bevfusion2,lss,bevdet,beverse}.
Other methods initialize queries in the 3D space and exploit deformable cross-attention to adaptively aggregate information from 2D images~\cite{bevformer,polarformer,tpvformer}.
Some works further design better positional embedding strategies~\cite{petr}, 3D representations~\cite{tpvformer}, or task heads~\cite{beverse} to further improve the perception performance or efficiency.
In this paper, we adopt conventional simple designs for 3D perception and focus on motion prediction and planning.

\textbf{Prediction.}  %
Accurate motion prediction for traffic participants is the key to the following motion planning of the ego vehicle.
Conventional methods utilized ground-truth agent history and HD map information as inputs and focused on the prediction of future agent trajectories~\cite{vaswani2017attention,chai2019multipath, phan2020covernet}.
One straightforward way is to draw agent paths and HD maps on a BEV image and use convolutional neural networks to process them and output motion prediction results~\cite{chai2019multipath, phan2020covernet}.
Further methods employed vectors or tokens to represent separate agents or map elements~\cite{liang2020lanegcn,vaswani2017attention, liu2021mmtrans, ngiam2021scenetrans}.
They then leverage the reasoning ability of graph neural networks~\cite{liang2020lanegcn} and transformers to infer future motions considering the interaction between agents and map elements.
The increase of hardware capacity promotes the emergence of end-to-end motion prediction methods~\cite{hu2021fiery, beverse, gu2022vip3d, jiang2022pip}, which jointly perform perception and prediction to get free of offline HD maps.
Despite being very challenging, recent end-to-end methods have demonstrated promising performance in this more practical setting~\cite{hu2021fiery, beverse, gu2022vip3d, jiang2022pip}.
They usually adopt the attention mechanism to incorporate agent and map information and leverage temporal networks (e.g., gated recurrent units~\cite{hu2021fiery}) to predict future states.
However, most existing methods directly decode trajectories from latent feature and ignores the structural nature of realistic trajectories (e.g., most of them are straight lines).
Differently, we learn a variational autoencoder from ground-truth trajectories to model the trajectory prior in a latent structural space and sample instances in this space for inference.

\begin{figure*}[t]
\centering
\includegraphics[width=\textwidth]{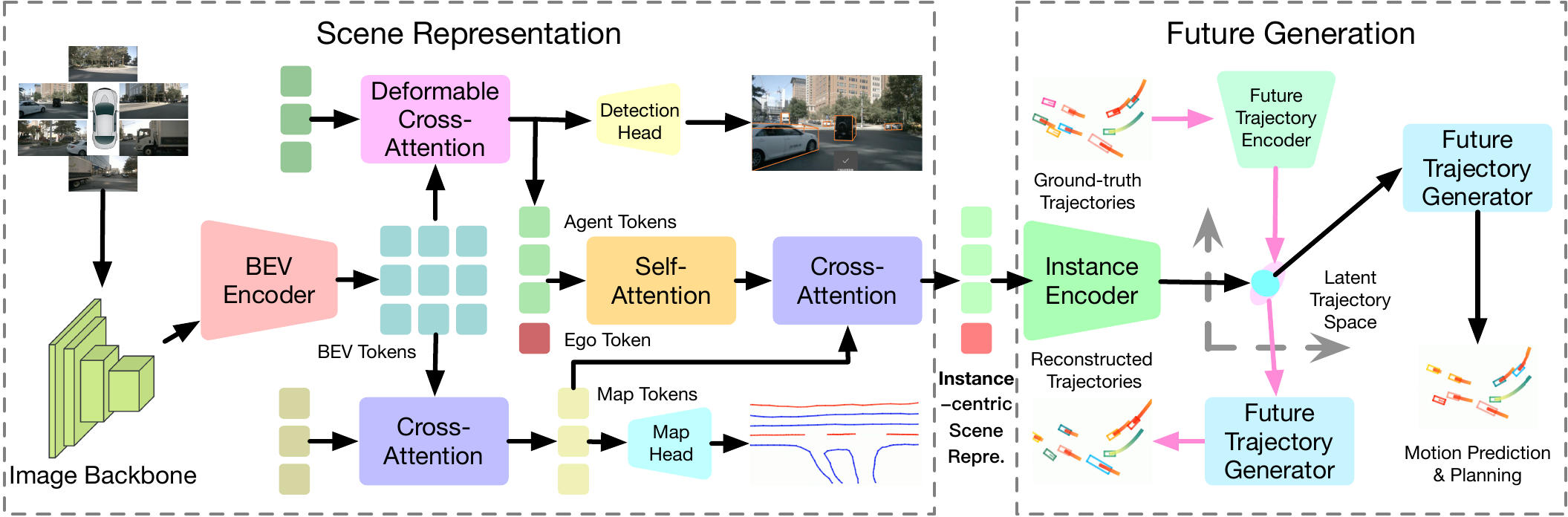}
\vspace{-7mm}
\caption{\textbf{Framework of our generative end-to-end autonomous driving.}
Given surrounding images as inputs, we employ an image backbone to extract multi-scale features and then use a BEV encoder to obtain BEV tokens.
We then use cross-attention and deformable cross-attention to transform BEV tokens into map and agent tokens, respectively.
With an additional ego token, we use self-attention to enable ego-agent interactions and cross-attention to further incorporate map information to obtain the instance-centric scene representation.
We map this representation to a structural latent trajectory space which is jointly learned using ground-truth future trajectories.
Finally, we employ a future trajectory generator to produce future trajectories to simultaneously complete motion prediction and planning.
}
\label{fig:framework}
\vspace{-5mm}
\end{figure*}

\textbf{Planning.}  
Planning is the ultimate goal for the first stage of autonomous driving.
Despite the mature development of rule-based planners~\cite{treiber2000idm,bouchard2022rule,Dauner2023CORL}, learning-based planners~\cite{scheel2022urban, cheng2022mpnp, pini2023safepathnet} are receiving increasing attention due to their great potential to benefit from large-scale driving data and compatibility with end-to-end autonomous driving methods.
Most existing end-to-end planning methods follow a pipeline of perception, prediction, and planning~\cite{hu2023uniad,hu2022stp3,vad,ppad}.
For example, ST-P3~\cite{hu2022stp3} progressively employs a map perception, a BEV occupancy prediction, and a trajectory planning module to obtain future ego movements from surrounding cameras. 
UniAD~\cite{hu2023uniad} further extends ST-P3 with additional detection, tracking, and motion prediction modules to improve the robustness of the system. 
VAD~\cite{vad} simplifies UniAD with vectorized scene representation and only map, motion, and planning modules for end-to-end driving, which achieves state-of-the-art planning performance with better efficiency.
However, the serial design of prediction and planning ignores the effect of future ego movements on the agent motion prediction.
It also lacks modeling of the uncertain nature of motion prediction and planning.
To address this, GenAD models autonomous driving in a generative framework and simultaneously generates the future trajectories for the ego vehicle and other agents in a learned probabilistic latent space.

\section{Proposed Approach}

This section presents our generative framework of vision-based end-to-end autonomous driving, as shown in Figure~\ref{fig:framework}.
We first introduce an instance-centric scene representation which incorporates high-order map-ego-agent interactions to enable comprehensive yet compact scene descriptions (Sec.~\ref{method sub: 1}).
We then elaborate on the learning of a latent embedding space to model realistic trajectories as prior (Sec.~\ref{method sub: 2}) and the generation of future motion in this learned latent space (Sec.~\ref{method sub: 3}).
At last, we detail the training and inference of the \textbf{Gen}erative end-to-end \textbf{A}utonomous \textbf{D}riving (GenAD) framework(Sec.~\ref{method sub: 4}).

\subsection{Instance-Centric Scene Representation} \label{method sub: 1}

The goal of end-to-end autonomous driving can be formulated as obtaining a planned $f$-frame future trajectory $\mathbf{T}(T,f) = \{ \mathbf{w}^{T+1}, \mathbf{w}^{T+2}, \cdots, \mathbf{w}^{T+f} \}$ for the ego vehicle given the current and past $p$-frame sensor inputs $\mathbf{S} = \{ \mathbf{s}^{T}, \mathbf{s}^{T-1}, \cdots, \mathbf{s}^{T-p} \}$ and trajectory $\mathbf{T}(T-p,p+1) = \{ \mathbf{w}^{T}, \mathbf{w}^{T-1}, \cdots, \mathbf{w}^{T-p} \}$.
\begin{equation}\label{eqn: AD_model}
\begin{aligned}
   \mathbf{T}(T-p,p+1), \mathbf{S} \rightarrow \mathbf{T}(T,f),
\end{aligned}
\end{equation}
where $\mathbf{T}(T,f)$ denotes a $f$-frame trajectory starting from the $T$-th frame, $\mathbf{w}^{t}$ denotes the waypoint at the $t$-th frame, and $\mathbf{s}^{t}$ denotes the sensor input at $t$-th frame.

The first step of end-to-end autonomous driving is to perceive the sensor inputs to obtain high-level descriptions of the surrounding scene.
These descriptions usually include semantic map~\cite{liao2022maptr,liu2022vectormapnet} and instance bounding box~\cite{bevformer,bevdepth}.
To achieve this, we follow a conventional vision-centric perception pipeline to extract bird's eye view (BEV) $\mathbf{B} \in \mathbb{R}^{H \times W \times C}$ features first and then build on them to refine map and bounding box features. 

\textbf{Image to BEV.} 
We basically follow BEVFormer~\cite{bevformer} to obtain the BEV features.
Specifically, we use a convolutional neural network~\cite{he2016deep} and feature pyramid network~\cite{lin2017feature} to obtain multi-scale image features $\mathbf{F}$ from camera inputs $\mathbf{s}$.
We then initialize $H \times W$ BEV tokens $\mathbf{B}_0$ as queries and use deformable cross-attention~\cite{zhu2020deformable} to transfer information from the multi-scale image feature $\mathbf{F}$:
\begin{equation}\label{eqn: BEV_feature}
\begin{aligned}
   \mathbf{B} = \text{DA}(\mathbf{B}_0, \mathbf{F}, \mathbf{F}),
\end{aligned}
\end{equation}
where $\textbf{DA} (\mathbf{Q},\mathbf{K},\mathbf{V})$ denotes the deformable attention block consisting of interleaved self-attention and deformable cross-attention layers using $\mathbf{Q}$, $\mathbf{K}$, and $\mathbf{V}$ as queries, keys, and values, respectively.
We then align the BEV features from the $p$ past frames into the current coordinate system and concatenate them as the final BEV features $\mathbf{B}$.

\textbf{BEV to map.} 
As semantic map elements are usually sparse in the BEV space, we follow a similar concept~\cite{jiang2022pip,vad} and use map tokens $\mathbf{M} \in \mathbb{R}^{N_{m}*C}$ to represent semantic maps.
Each map token $\mathbf{m} \in \mathbf{M}$ can be decoded to a set of points in the BEV space by a map decoder $d_{m}$ representing a category of map elements and their corresponding positions.
Following VAD~\cite{vad}, we consider three categories of map elements (i.e., lane divider, road boundary, and pedestrian crossing).
We use the global cross-attention mechanism to update learnable initialized queries $\mathbf{M}_0$ from BEV tokens $\mathbf{B}$:
\begin{equation}\label{eqn: map_tokens}
\begin{aligned}
   \mathbf{M} = \text{CA}(\mathbf{M}_0, \mathbf{B}, \mathbf{B}),
\end{aligned}
\end{equation}
where $\textbf{CA} (\mathbf{Q},\mathbf{K},\mathbf{V})$ denotes the cross-attention block composed of interleaved self-attention and cross-attention layers using $\mathbf{Q}$, $\mathbf{K}$, and $ \mathbf{V}$ as queries, keys, and values, respectively.

\textbf{BEV to agent.}
Similar to the representation of semantic maps, we adopt a set of agent tokens $\mathbf{A}$ to represent the 3D position of each instance in the surroundings. 
We use deformable cross-attention to obtain the updated agent tokens $\mathbf{A}$ from the BEV tokens $\mathbf{B}$: 
\begin{equation}\label{eqn: agent_tokens}
\begin{aligned}
   \mathbf{A} = \text{DA}(\mathbf{A}_0, \mathbf{B}, \mathbf{B}),
\end{aligned}
\end{equation}
where $\mathbf{A}_0$ are learnable tokens as initialization.

Having obtained the agent tokens $\mathbf{A}$, we employ a 3D object detection head $d_a$ to decode the position, orientation, and category information from each agent token $\mathbf{a}$.

\textbf{Instance-centric scene representation.}
As prediction and planning mainly focus on the instances of agents and ego vehicles, respectively, we propose an instance-centric scene representation to comprehensively and efficiently represent the autonomous driving scenario.
We first add an ego token $\mathbf{e}$ to the learned agent tokens $\mathbf{A}$ to construct a set of instance tokens $\mathbf{I} = \text{concat}({\mathbf{e}, \mathbf{A}})$.	

Existing methods~\cite{hu2023uniad,hu2022stp3,vad} usually perform motion prediction and planning in a serial manner, which ignores the effect of future ego movements on the agents.
For example, the lane shift of the ego car would possibly affect the action of rear cars, rendering the motion prediction results inaccurate.
Differently, we enable high-order interactions between the ego vehicle and other agents by performing self-attention on the instance tokens:
\begin{equation}\label{eqn: instance_tokens}
\begin{aligned}
   \mathbf{I} \leftarrow \text{SA}(\mathbf{I}, \mathbf{I}, \mathbf{I}),
\end{aligned}
\end{equation}
where $\textbf{SA} (\mathbf{Q},\mathbf{K},\mathbf{V})$ denotes the self-attention block composed of self-attention layers using $\mathbf{Q}$, $\mathbf{K}$, and $\mathbf{V}$ as queries, keys, and values, respectively.

Furthermore, to perform accurate prediction and planning, both the agents and the ego vehicle need to be aware of the semantic map information.
We thus employ cross-attention between the updated instance tokens and the learned map tokens to obtain map-aware instance-centric scene representations:
\begin{equation}\label{eqn: map_aware}
\begin{aligned}
   \mathbf{I} \leftarrow \text{CA}(\mathbf{I}, \mathbf{M}, \mathbf{M}).
\end{aligned}
\end{equation}

The learned instance tokens $\mathbf{I}$ incorporate high-order agent-ego interactions and are aware of the learned semantic maps, which are compact yet contain all the necessary map and instance information to perform motion prediction and trajectory planning.

\subsection{Trajectory Prior Modeling} \label{method sub: 2}
We find that the objectives of motion prediction of other agents and planning of the ego vehicle share the same output space and are essentially the same.
They both aim to produce a high-quality realistic trajectory of the concerned instance, given semantic maps and interactions with other agents.
The goal for the proposed GenAD can be then formulated as inferring the future trajectory $\mathbf{T}$ given the map-aware instance-centric scene representations $\mathbf{I}$.

Different from existing methods which directly output the trajectory using a simple decoder, we model it as a trajectory generation problem $\mathbf{T} \sim p(\mathbf{T}|\mathbf{I}) $ considering its uncertain nature.

\begin{figure}[t]
\centering
\includegraphics[width=0.475\textwidth]{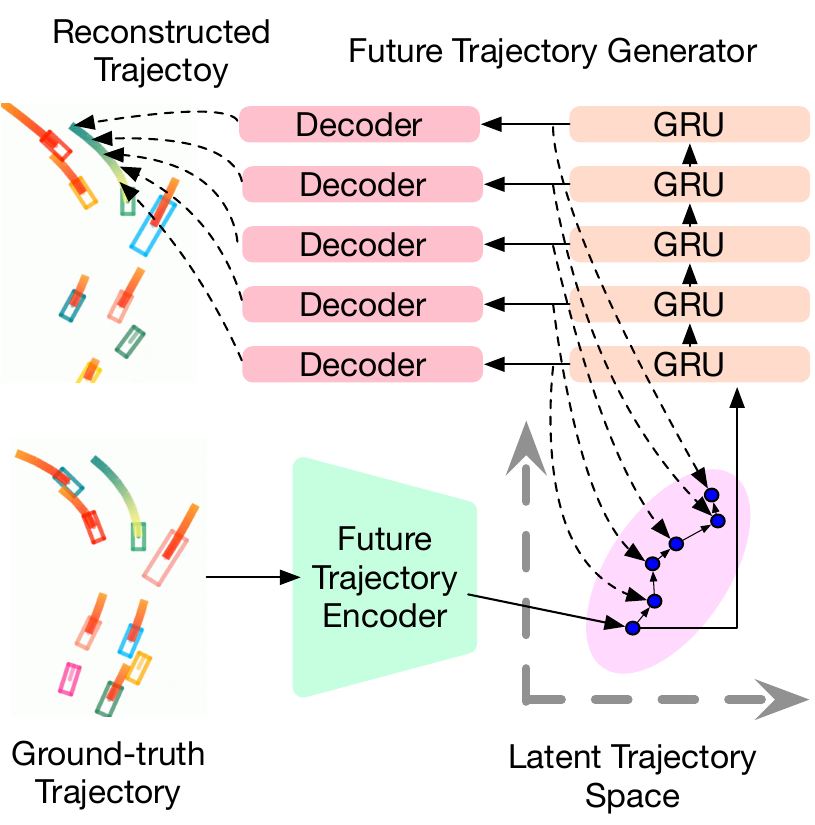}
\vspace{-7mm}
\caption{\textbf{Illustration of the proposed trajectory prior modeling and future generation.}
We use a future trajectory encoder to map ground-truth trajectories to a latent trajectory space, where we use the Gaussian distribution to model the trajectory uncertainty.
We then employ a gate recurrent unit (GRU) to progressively predict the next future in the latent space and use a decoder to obtain explicit trajectories.
}
\label{fig:generator}
\vspace{-6mm}
\end{figure}

The trajectories of both the ego vehicle and other agents are highly structured (e.g., continuous) and follow certain patterns.
For example, most of the trajectories are straight lines as the vehicle driving at a constant speed, and some of them are curved lines with a near-constant curvature when the vehicle turning right or left.
Only in very rare cases will the trajectories be zig-zagging.
Considering this, we adopt a variational autoencoder (VAE)~\cite{vae} architecture to learn a latent space $\mathbb{Z}$ to model this trajectory prior.
Specifically, we employ a ground-truth trajectory encoder $e_{f}$ to model $p(\mathbf{z}|\mathbf{T}(T,f))$, which maps the future trajectory $\mathbf{T}(T,f)$ to a diagonal Gaussian distribution on the latent space $\mathbb{Z}$.
The encoder $e_{f}$ outputs two vectors $\mathbf{\mu}_f$ and $\mathbf{\sigma}_f$ representing the mean and variance of the Gaussian distribution:
\begin{equation}\label{eqn: prior}
\begin{aligned}
   p(\mathbf{z}|\mathbf{T}(T,f)) \sim N(\mathbf{\mu}_f,\mathbf{\sigma}_f),
\end{aligned}
\end{equation}
where $N(\mathbf{\mu},\mathbf{\sigma}^2)$ denotes a Gaussian distribution with a mean of $\mathbf{\mu}$ and standard deviation of $\mathbf{\sigma}$.

The learned distribution $p(\mathbf{z}|\mathbf{T}(T,f))$ contains the prior of the ground-truth trajectories, which can be leveraged to improve the authenticity of motion prediction and planning for traffic agents and ego vehicles.

\subsection{Latent Future Trajectory Generation} \label{method sub: 3}
Having obtained the latent distribution of the future trajectories as prior, we need to explicitly decode them from the latent trajectory space $\mathbb{Z}$.

While a straightforward way is to directly use an MLP-based decoder to output trajectory points in the BEV space to model $p(\mathbf{T}(T,f)|\mathbf{z})$, it fails to model the temporal evolution of the traffic agents and ego vehicle.
To consider the temporal relations of instances at different time stamps, we factorized the joint distribution $p(\mathbf{T}(T,f)|\mathbf{z})$ as follows:
\begin{equation}\label{eqn: plannings}
\begin{aligned}
   p(\mathbf{T}(T,f)|\mathbf{z}^T) = p(\mathbf{w}^{T+1}|\mathbf{z}^T) \cdot p(\mathbf{w}^{T+2}|\mathbf{w}^{T+1},\mathbf{z}^T) \\
   \cdots p(\mathbf{w}^{T+f}|\mathbf{w}^{T+1},\cdots, \mathbf{w}^{T+f-1}, \mathbf{z}^T).
\end{aligned}
\end{equation}

We sample a vector from the distribution $N(\mathbf{\mu}_f,\mathbf{\sigma}_f)$ as the latent state at the current time stamp $\mathbf{z}^T$.
Instead of decoding the whole trajectory at once, we adopt a simple MLP-based decoder $d_w$ to decode a waypoint $\mathbf{w} = d_w (\mathbf{z})$ from the latent space $\mathbf{Z}$, i.e., we instantiate $p(\mathbf{w}^{T+1}|\mathbf{z})$ with $\mathbf{w} = d_w (\mathbf{z})$.

We then adopt a gated recurrent unit (GRU)~\cite{gru} as the future trajectory generator to model the temporal evolutions of instances.
Specifically, the GRU model $g$ takes as inputs the current latent representation $\mathbf{z}_t$ and transforms it into the next state $g(\mathbf{z}_t) = \mathbf{z}_{t+1}$.
We can then decode the waypoint $\mathbf{w}^{t+1}$ at the $(t+1)$-th time stamp using the waypoint decoder $\mathbf{w}^{t+1} = d_w(\mathbf{z}_{t+1})$, i.e., we model $p(\mathbf{w}^{t+1}|\mathbf{w}^{T+1},\cdots, \mathbf{w}^{t}, \mathbf{z})$ with $d_w(g(\mathbf{z}_t))$.

Compared with a single decoder that directly outputs the whole trajectory, the waypoint decoder performs a simpler task of only decoding a position in the BEV space and the GRU module models the movement of agents in the latent space $\mathbf{Z}$.
The produced trajectory is thus more realistic and authentic considering the prior knowledge in this learned structured latent space. 
We illustrate the proposed trajectory prior modeling and latent future trajectory generation in Figure~\ref{fig:generator}.

\subsection{Generative End-to-End Autonomous Driving} \label{method sub: 4}
In this subsection, we present the overall architecture of the proposed GenAD framework for vision-centric end-to-end autonomous driving.
Given surrounding camera signals $\mathbf{s}$ as inputs, we first employ an image backbone to extract multi-scale image features $F$ and then use deformable attention to transform them into the BEV space.
We align the BEV features from the past $p$ frames to the current ego-coordinate to obtain the final BEV feature $\mathbf{B}$.
We perform global cross-attention and deformable attention to refine a set of map tokens $\mathbf{M}$ and agent tokens $\mathbf{A}$, respectively.
To model the high-order interactions between traffic agents and the ego vehicle, we combine agent tokens with an ego token and perform self-attention among them to construct a set of instance tokens $\mathbf{I}$.
We also use cross-attention to inject semantic map information into the instance tokens $\mathbf{I}$ to facilitate further prediction and planning.

As realistic trajectories are highly structured, we learn a VAE module to model the trajectory prior and adopt a  generative framework for both motion prediction and planning.
We learn an encoder $e_t$ to map ground-truth trajectories to the structural space $\mathbf{Z}$ as Gaussian distributions.
We then employ a GRU-based future trajectory generator $g$ to model the temporal evolutions of instances in the latent space $\mathbf{Z}$ and use a simple MLP-based decoder $d_w$ to decode waypoints from latent representations.
We can finally reconstruct the trajectories $\hat{\mathbf{T}}_a$ and $\hat{\mathbf{T}}_e$ for traffic agents and the ego vehicle by integrating the decoded waypoint of each time stamp.
For training, we additionally use a class decoder $d_c$ to predict the category for each agent $\hat{\mathbf{c}}_a$.
To learn the future trajectory encoder $e_f$, the future trajectory generator $g$, the waypoint decoder $d_w$, and the class decoder $d_c$, we follow VAD~\cite{vad} to impose trajectory losses on the reconstructed and ground-truth trajectories for both traffic agents and the ego vehicle:
\begin{equation}\label{eqn: obj_prior}
\begin{aligned}
   J_{prior} = L_{tra}(\hat{\mathbf{T}}_e, \mathbf{T}_e) + \frac{1}{N_a} L_{tra}(\hat{\mathbf{T}}_a, \mathbf{T}_a) \\
   + \lambda_{c} L_{focal} (\hat{\mathbf{C}}_a, \mathbf{C}_a),
\end{aligned}
\end{equation}
where $||\cdot||_1$ denotes the L1 norm, $N_a$ is the number of agents, and $L_{focal}$ represents the focal loss to constrain the predicted agent class. 
$L_{tra}$ denotes the trajectory losses~\cite{vad} including the L1 discrepancy, ego-agent collision constraint, ego-boundary overstepping constraint, and the ego-Lane directional constraint.
$\hat{\mathbf{C}}_a$ and $\mathbf{C}_a$ represent the predicted and ground-truth classes for all the agents, respectively.

\begin{table*}[t]
\setlength{\tabcolsep}{0.009\linewidth}
\caption{\textbf{Comparisons with state-of-the-art methods in motion planning performance on the nuScenes~\cite{caesar2020nuscenes} val dataset.} 
$\dagger$ represents that the metrics are computed with an average of all the predicted frames.
$^\dagger$ denotes FPS measured with the same environment on our machine with a single RTX 3090 GPU.
}
\vspace{-3mm}
\begin{tabular}{l|cc|cccc|cccc|c}
\toprule
\multirow{2}{*}{Method} & \multirow{2}{*}{Input} & \multirow{2}{*}{Aux. Sup.} &
\multicolumn{4}{c|}{L2 (m) $\downarrow$} & 
\multicolumn{4}{c|}{Collision Rate (\%) $\downarrow$} &
\cellcolor{gray!30} \\
&& & 1s & 2s & 3s & \cellcolor{gray!30}Avg. & 1s & 2s & 3s & \cellcolor{gray!30}Avg. & \cellcolor{gray!30}\multirow{-2}*{FPS} \\
\midrule
IL~\cite{ratliff2006maximum} & LiDAR & None  & 0.44 & 1.15 & 2.47 & \cellcolor{gray!30}1.35 & 0.08 & 0.27 & 1.95 & \cellcolor{gray!30}0.77 & \cellcolor{gray!30}- \\
NMP~\cite{zeng2019nmp} & LiDAR & Box \& Motion & 0.53 & 1.25 & 2.67 & \cellcolor{gray!30}1.48 & 0.04 & 0.12 & 0.87 & \cellcolor{gray!30}0.34 & \cellcolor{gray!30}- \\
FF~\cite{hu2021ff} & LiDAR & Freespace  & 0.55 & 1.20 & 2.54 & \cellcolor{gray!30}1.43 & 0.06 & 0.17 & 1.07 & \cellcolor{gray!30}0.43 & \cellcolor{gray!30}- \\
EO~\cite{khurana2022eo} & LiDAR & Freespace  & 0.67 & 1.36 & 2.78 & \cellcolor{gray!30}1.60 & 0.04 & 0.09 & 0.88 & \cellcolor{gray!30}0.33 & \cellcolor{gray!30}- \\
\midrule
\midrule
ST-P3~\cite{hu2022stp3} & Camera & Map \& Box \& Depth & 1.33 & 2.11 & 2.90 & \cellcolor{gray!30}2.11 & 0.23 & 0.62 & 1.27 & \cellcolor{gray!30}0.71 &  \cellcolor{gray!30}1.6 \\
UniAD~\cite{hu2023uniad} & Camera & { \footnotesize Map \& Box \& Motion \& Tracklets \& Occ}  & \underline{0.48} & \underline{0.96} & \underline{1.65} & \cellcolor{gray!30}\underline{1.03} & \underline{0.05} & \textbf{0.17} & \textbf{0.71} & \cellcolor{gray!30}\textbf{0.31} & \cellcolor{gray!30}1.8 \\
OccNet~\cite{tong2023scene} & Camera & 3D-Occ \& Map \& Box & 1.29 & 2.13 & 2.99 & \cellcolor{gray!30}2.14 & 0.21 & 0.59 & 1.37 & \cellcolor{gray!30}0.72  & \cellcolor{gray!30}2.6 \\
VAD-Tiny~\cite{vad}  & Camera & Map \& Box \& Motion  & 0.60 & 1.23 & 2.06 & \cellcolor{gray!30}1.30 & 0.31 & 0.53 & 1.33 & \cellcolor{gray!30}0.72 & \cellcolor{gray!30}\textbf{6.9}$^\ddagger$ \\
VAD-Base~\cite{vad} & Camera & Map \& Box \& Motion & 0.54 & 1.15 & 1.98 & \cellcolor{gray!30}1.22 & \textbf{0.04} & 0.39 & 1.17 & \cellcolor{gray!30}0.53 & \cellcolor{gray!30}3.6$^\ddagger$ \\
\midrule
GenAD & Camera & Map \& Box \& Motion & \textbf{0.36} & \textbf{0.83} & \textbf{1.55} & \cellcolor{gray!30}\textbf{0.91} & 0.06 & \underline{0.23} & \underline{1.00} & \cellcolor{gray!30}\underline{0.43}  & \cellcolor{gray!30}\underline{6.7}$^\ddagger$ \\
\midrule
\midrule
\color{gray}VAD-Tiny$^\dagger$~\cite{vad}  & \color{gray}Camera & \color{gray}Map \& Box \& Motion  & \color{gray}0.46 & \color{gray}0.76 & \color{gray}1.12 & \color{gray}\cellcolor{gray!30}0.78 & \color{gray}0.21 & \color{gray}0.35 & \color{gray}0.58 & \color{gray}\cellcolor{gray!30}0.38 & \color{gray}\cellcolor{gray!30}\textbf{6.9}$^\ddagger$ \\
\color{gray}VAD-Base$^\dagger$~\cite{vad} & \color{gray}Camera & \color{gray}Map \& Box \& Motion & \color{gray}0.41 & \color{gray}0.70 & \color{gray}1.05 & \color{gray}\cellcolor{gray!30}0.72 & \color{gray}\textbf{0.07} & \color{gray}0.17 & \color{gray}0.41 & \color{gray}\cellcolor{gray!30}0.22 & \color{gray}\cellcolor{gray!30}3.6$^\ddagger$ \\
\midrule
\color{gray}GenAD$^\dagger$ & \color{gray}Camera & \color{gray}Map \& Box \& Motion & \color{gray}\textbf{0.28} & \color{gray}\textbf{0.49} & \color{gray}\textbf{0.78} & \color{gray}\cellcolor{gray!30}\textbf{0.52} & \color{gray}0.08 & \color{gray}\textbf{0.14} & \color{gray}\textbf{0.34} & \color{gray}\cellcolor{gray!30}\textbf{0.19}  & \color{gray}\cellcolor{gray!30}6.7 \\
\bottomrule
\end{tabular}%
\label{tab:sota-plan}
\vspace{-4mm}
\end{table*}

To infer the future trajectory for traffic agents and the ego vehicle from the instance tokens $\mathbf{I}$, we use an instance encoder $e_i$ to map each instance token to the latent space $\mathbb{Z}$.
The encoder $e_i$ similarly outputs a mean vector $\mu_i$ and variance vector $\sigma_i$ to parameterize a diagonal Gaussian distribution:
\begin{equation}\label{eqn: instance}
\begin{aligned}
   p(\mathbf{z}|\mathbf{I}) \sim N(\mathbf{\mu}_i,\mathbf{\sigma}_i).
\end{aligned}
\end{equation}
With the learned latent trajectory space to model realistic trajectory prior, the motion prediction and planning can be unified and formulated as a distribution matching problem between the instance distribution $p(\mathbf{z}|\mathbf{I})$ and the ground-truth distribution $p(\mathbf{z}|\mathbf{T}(T,f))$.
We impose the Kullback-Leibler divergence loss to enforce distribution matching:
\begin{equation}\label{eqn: instance}
\begin{aligned}
   J_{plan} = D_{KL} (p(\mathbf{z}|\mathbf{I}), p(\mathbf{z}|\mathbf{T}(T,f))),
\end{aligned}
\end{equation}
where $D_{KL}$ denotes the Kullback-Leibler divergence.

Additionally, we use two auxiliary tasks to train the proposed GenAD model: map segmentation and 3D object detection.
We use a map decoder $d_m$ on the map tokens $\mathbf{M}$ and an object decoder $d_o$ on the agent tokens $\mathbf{A}$ to obtain the predicted maps and 3D object detection results. 
We follow the task decoder design of VAD~\cite{vad} and employ bipartite matching for ground truth matching.
We then impose semantic map loss~\cite{beverse} $J_{map}$ and 3D object detection loss~\cite{bevformer} $J_{det}$ on them to train the network.

The overall training objective of our GenAD framework can be formulated as:
\begin{equation}\label{eqn: overall_obj}
\begin{aligned}
   J_{GenAD} = J_{prior} + \lambda_{plan} J_{plan} + \lambda_{map} J_{map} + \lambda_{det} J_{det},
\end{aligned}
\end{equation}
where $\lambda_{plan}$, $\lambda_{map}$, and $\lambda_{det}$ are balance factors. 
The proposed GenAD can be trained efficiently in an end-to-end manner.
For inference, we discard the future trajectory encoder $e_{f}$ and sample a latent state following the instance distribution $p(\mathbf{z}|\mathbf{I})$ as input for the trajectory generator $g$ and waypoint decoder $d_w$.
Our GenAD models end-to-end autonomous driving as a generative problem and performs future prediction and planning in a structured latent space, which considers the prior of realistic trajectories to produce high-quality trajectory prediction and planning.

\begin{table*}[t]
\setlength{\tabcolsep}{0.007\linewidth}
\caption{\textbf{Comparisons of perception, prediction, and planning performance.} 
$^\dagger$ denotes FPS measured with the same environment on our machine with a single RTX 3090 GPU card.
}
\vspace{-3mm}
\begin{tabular}{l|c|ccc|cc|cc|c}
\toprule
\multirow{2}{*}{Method} &
\multicolumn{1}{c|}{Detection} & \multicolumn{3}{c|}{Map Segmentation} & \multicolumn{2}{c|}{Motion Prediction} & \multicolumn{2}{c|}{Planning} & \multirow{2}{*}{FPS}  \\
 & mAP $\uparrow$ & mAP@0.5 $\uparrow$ & mAP@1.0 $\uparrow$ & mAP@1.5 $\uparrow$ & EPA (car) $\uparrow$ & EPA (ped.)$\uparrow$ & Avg. L2 $\downarrow$ & Avg. CR. $\downarrow$\\
\midrule
VAD~\cite{vad} & 0.27 & 0.15 & 0.44 & 0.61 & 0.56 & 0.29 & 1.30 & 0.72	 & \textbf{6.9}$^\dagger$  \\
GenAD & \textbf{0.29} & \textbf{0.24} & \textbf{0.56} & \textbf{0.71} & \textbf{0.59} & \textbf{0.34} & \textbf{0.91} & \textbf{0.43} & 6.7$^\dagger$  \\
\bottomrule
\end{tabular}%
\label{tab:perception}
\vspace{-4mm}
\end{table*}

\begin{table}[t]
\setlength{\tabcolsep}{0.01\linewidth}
\caption{\textbf{Effect of the instance-centric scene representation.} 
E $\rightarrow$ A represents the proposed ego-to-agent interaction to obtain the instance-centric scene representation.
}
\vspace{-3mm}
\begin{tabular}{l|cccc|cccc}
\toprule
\multirow{2}{*}{Setting} &
\multicolumn{4}{c|}{L2 (m) $\downarrow$} & 
\multicolumn{4}{c}{Collision Rate (\%) $\downarrow$}  \\
 & 1s & 2s & 3s & \cellcolor{gray!30}Avg. & 1s & 2s & 3s & \cellcolor{gray!30}Avg.  \\
\midrule
VAD~\cite{vad} & 0.60 & 1.23 & 2.06 & \cellcolor{gray!30}1.30 & 0.31 & 0.53 & 1.33 & \cellcolor{gray!30}0.72	  \\
w/ E $\rightarrow$ A  & \textbf{0.39} & \textbf{0.99} & \textbf{1.91}  & \cellcolor{gray!30}\textbf{1.10} & \textbf{0.14} & \textbf{0.39} & \textbf{1.35} & \cellcolor{gray!30}\textbf{0.63}  \\
\hline
GenAD & \textbf{0.36} & \textbf{0.83} & \textbf{1.55} & \cellcolor{gray!30}\textbf{0.91} & \textbf{0.06} & \textbf{0.23} & \textbf{1.00} & \cellcolor{gray!30}\textbf{0.43}  \\
w/o  E $\rightarrow$ A  & 0.40 & 0.93 & 1.74  & \cellcolor{gray!30}1.02 & 0.70 & 0.98 & 1.97 & \cellcolor{gray!30}1.22  \\
\bottomrule
\end{tabular}%
\label{tab:repre}
\vspace{-4mm}
\end{table}

\section{Experiments}

\subsection{Datasets}
We conducted extensive experiments on the widely adopted nuScenes~\cite{caesar2020nuscenes} dataset to evaluate the proposed GenAD framework for autonomous driving.
The nuScenes dataset is composed of 1000 driving scenes, where each scene provides RGB and LiDAR video of 20 seconds.
The ego vehicle is equipped with 6 surrounding cameras with $360^{\circ}$ horizontal FOV and a 32-beam LiDAR sensor.
nuScenes provides semantic map and 3D object detection annotations for keyframes at 2Hz.
It includes 1.4M 3D bounding boxes of objects from 23 categories.
We partitioned the dataset into 700, 150, and 150 scenes for training, validation, and testing, respectively, following the official instructions~\cite{caesar2020nuscenes}.

\subsection{Evaluation Metrics}
Following existing end-to-end autonomous driving methods~\cite{hu2022stp3, hu2023uniad}, we use the L2 displacement error and collision rate to measure the quality of planning results.
The L2 displacement error measures the L2 distance between the planned trajectory and the ground-truth trajectory.
The collision rate measures how often the self-driving vehicle collapses with other traffic participants following the planned trajectory.
By default, we take as inputs 2s history (i.e., 5 frames) and evaluate the planning performance at the 1s, 2s, and 3s future.

\subsection{Implementation Details}
We adopted ResNet50~\cite{he2016resnet} as the backbone network to extract image features.
We take as input images with a resolution of $640 \times 360$ and use a $200 \times 200$ BEV representation to perceive the surrounding scene.
For fair comparisons, we basically use the same hyperparameters as VAD-tiny\cite{vad}.
We fixed the number of BEV tokens, map tokens, and agent tokens to $100 \times 100$, 100, and 300, respectively.
Each map token contains 20 point tokens to represent a map point in the BEV space.
We set the hidden dimension of each BEV, point, agent, ego, and instance token to 256.
We used a latent space with a dimension of 512 to model trajectory prior and also set the hidden dimension of the GRU to 512.
We employed 3 layers for each attention block.

For training, we set the loss balance factors to 1 and use the AdamW~\cite{loshchilov2017adamw} optimizer with a cosine learning rate scheduler~\cite{loshchilov2016cosineanneal}.
We set the initial learning rate to $2\times{10}^{-4}$ and a weight decay of 0.01.
By default, we trained our GenAD for 60 epochs with 8 NVIDIA RTX 3090 GPUs and adopted a total batch size of 8.

\subsection{Results and Analysis}

\textbf{Main results.}
We compared GenAD with state-of-the-art end-to-end autonomous driving methods in Table~\ref{tab:sota-plan}.
We use bold and underlined numbers to represent the best and second-best results, respectively.
We see that our GenAD achieves the best L2 errors among all the methods with an efficient inference speed.
Though UniAD~\cite{hu2023uniad} outperforms our method with respect to the collision rates, it employs additional supervision signals during training such as tracking and occupancy information, which has been verified to be vital to avoid collision~\cite{hu2023uniad}.
However, these labels in the 3D space are difficult to annotate, rendering it not trivial to use less labels to achieve competitive performance.
Our GenAD is also more efficient than UniAD, demonstrating a strong performance/speed trade-off.

\textbf{Perception and prediction performance.}
We further evaluated the perception and prediction performance of the proposed GenAD model and compared it with VAD-tiny~\cite{vad} with a similar model size, as shown in Table~\ref{tab:perception}.
We use mean average precision (mAP) to measure the 3D object detection performance, and mAP@0.5, mAP@1.0, and mAP@1.5 to evaluate the quality of predicted maps.
For motion prediction, we report the end-to-end prediction accuracy (EPA) for both cars and pedestrians, which is a more fair metric for end-to-end methods to avoid the influence of falsely detected agents.
For motion planning, we report the average L2 error and collision rate (CR) over 1s, 2s, and 3s.

We observe that GenAD outperforms VAD on all the tasks with a similar inference speed. 
Specifically, GenAD achieves better motion prediction performances by considering the influence of the ego vehicle on the other agents.
GenAD also demonstrates superior performance in 3D detection and map segmentation, showing a better consistency between perception, prediction, and planning.

\begin{table}[t]
\setlength{\tabcolsep}{0.009\linewidth}
\caption{\textbf{Effect of the generative framework for autonomous driving.} 
TPM and LFTG denote the trajectory prior modeling and latent future trajectory generation, respectively.
}
\vspace{-3mm}
\begin{tabular}{cc|cccc|cccc}
\toprule
\multirow{2}{*}{TPM} & \multirow{2}{*}{LFTG} &
\multicolumn{4}{c|}{L2 (m) $\downarrow$} & 
\multicolumn{4}{c}{Collision Rate (\%) $\downarrow$}  \\
& & 1s & 2s & 3s & \cellcolor{gray!30}Avg. & 1s & 2s & 3s & \cellcolor{gray!30}Avg.  \\
\midrule
$\times$ & $\times$ & 0.42 & 1.02 & 1.89 & \cellcolor{gray!30}1.11 & 0.06 & 0.43 & 1.60 & \cellcolor{gray!30}0.70	  \\
$\checkmark$ & $\times$ & 0.38 & 0.92 & 1.76  & \cellcolor{gray!30}1.02 & 0.23 & 0.37 & 1.19 & \cellcolor{gray!30}0.60  \\
$\times$ & $\checkmark$ & 0.39 & 0.92 & 1.74  & \cellcolor{gray!30}1.02 & 0.18 & 0.45 & 1.05 & \cellcolor{gray!30}0.56  \\
$\checkmark$ & $\checkmark$ & \textbf{0.36} & \textbf{0.83} & \textbf{1.55} & \cellcolor{gray!30}\textbf{0.91} & \textbf{0.06} & \textbf{0.23} & \textbf{1.00} & \cellcolor{gray!30}\textbf{0.43}  \\
\bottomrule
\end{tabular}%
\label{tab:generator}
\vspace{-4mm}
\end{table}

\begin{figure*}[t]
\centering
\includegraphics[width=\textwidth]{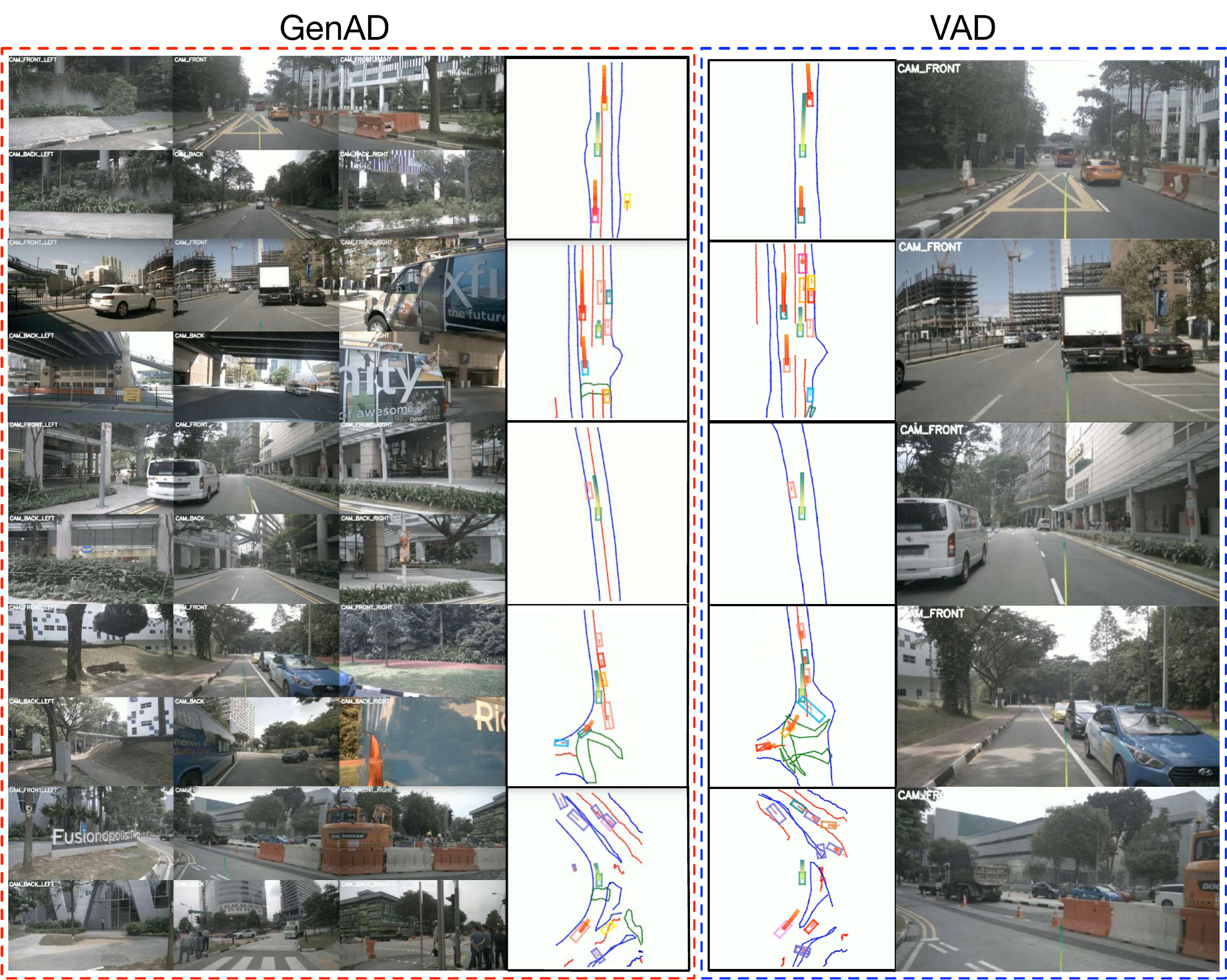}
\vspace{-7mm}
\caption{\textbf{Visualizations of the results of GenAD with comparisons with VAD~\cite{vad}.}
We provide perception, motion prediction, and planning results with surrounding camera inputs.
}
\label{fig:vis}
\vspace{-5mm}
\end{figure*}

\textbf{Effect of the instance-centric scene representation.}
We conducted an ablation study to analyze the effectiveness of the instance-centric scene representation, as shown in Table~\ref{tab:repre}.
We first added the ego-to-agent interaction with the proposed method to VAD-tiny~\cite{vad}, and observe a large improvement in both the L2 error and the collision rate.
We also removed the ego-to-agent interaction in our GenAD model by masking the self-attention matrix to dissect its effect.
We see that the collision rate performance drops greatly.
We think this is because without considering the high-order interactions between the ego car and the other agents, it becomes very difficult to learn the true underlying distributions of trajectories.

\textbf{Effect of the generative framework of autonomous driving.}
We also analyzed the designs of the proposed future trajectory generation model, which is composed of two modules: trajectory prior modeling (TPM) and latent future trajectory generation (LFTG).
When using TPM alone, we directly decode the entire trajectory from the latent space.
With only the LFTG module, we use a gated recurrent unit to gradually produce waypoints given the instance-centric scene representation.
We see that both modules are effective and improve the planning performance.
Combining the two modules further improves the performance by a large margin.
This verifies the importance of factorizing the joint distribution as in \eqref{eqn: plannings} to release the full potential of latent trajectory prior modeling.

\textbf{Visualizations.}
We provide a visualization of our GenAD model with comparisons with VAD-tiny~\cite{vad} with a similar model size, as shown in Figure~\ref{fig:vis}.
We visualize the map segmentation, detection, motion prediction, and planning results on a single image and provide the surrounding camera inputs as references.
We see that GenAD produces better and safer trajectories than VAD in various scenarios including going straight, overtaking, and turning.
For the challenging scenarios when multiple agents are involved in complex traffic scenes, our GenAD still demonstrates good results while VAD cannot safely move through.

\section{Conclusion}
In this paper, we have presented a generative end-to-end autonomous driving (GenAD) framework for better planning from vision inputs.
We have investigated the conventional serial design of perception, prediction, and planning for autonomous driving and proposed a generative framework to enable high-order ego-agent interactions and produce more accurate future trajectories with learned structural prior.
We have conducted extensive experiments on the widely adopted nuScenes dataset and demonstrated the state-of-the-art planning performance of the proposed GenAD.
In the future, it is interesting to explore other generative modeling methods such as generative adversarial networks or diffusion models for end-to-end autonomous driving.

{
    \small
    \bibliographystyle{ieeenat_fullname}
    \bibliography{main}
}

\end{document}